\documentclass[12pt,doublecolumn]{IEEEtran}
\usepackage{amsmath,amsthm,amsfonts,amssymb,bm,mathrsfs}
\usepackage{graphicx,subfigure}
\usepackage{multicol}
\usepackage[square, comma, sort&compress, numbers]{natbib}
\usepackage[colorlinks, linkcolor=black, citecolor=black, urlcolor=black]{hyperref}
\usepackage{mathrsfs}
\usepackage{algorithm}
\usepackage{algorithmic}
\usepackage{multirow,booktabs,threeparttable}
\usepackage{tablefootnote}
\usepackage{color}
\newcommand{\subparagraph}{}
\usepackage{verbatim}
\usepackage{caption}
\usepackage{url}
\usepackage[perpage,symbol*]{footmisc}
\usepackage{titlesec}
\theoremstyle{remark}

\theoremstyle{plain}
\newtheorem*{remark}{Remark}

\begin{document}

\title{Deep Learning with Long Short-Term Memory for Time Series Prediction}

\author{
Yuxiu Hua, Zhifeng Zhao, Rongpeng Li, Xianfu Chen, Zhiming Liu, 

and Honggang Zhang

\thanks{Y. Hua, Z. Zhao, R. Li and H. Zhang are with Zhejiang University, Hangzhou 310027, China, (email: \{21631087, zhaozf, lirongpeng, honggangzhang\}@zju.edu.cn).}

\thanks{X. Chen is with VTT Technical Research Centre of Finland, Oulu FI-90571, Finland (email: xianfu.chen@vtt.fi).}

\thanks{Z. Liu is with China Mobile Research Institute, Beijing 100053, China (email: liuzhiming@chinamobile.com).}
}

\maketitle

\begin{abstract}
Time series prediction can be generalized as a process that extracts useful information from historical records and then determines future values. Learning long-range dependencies that are embedded in time series is often an obstacle for most algorithms, whereas Long Short-Term Memory (LSTM) solutions, as a specific kind of scheme in deep learning, promise to effectively overcome the problem. In this article, we first give a brief introduction to the structure and forward propagation mechanism of the LSTM model. Then, aiming at reducing the considerable computing cost of LSTM, we put forward the Random Connectivity LSTM (RCLSTM) model and test it by predicting traffic and user mobility in telecommunication networks. Compared to LSTM, RCLSTM is formed via stochastic connectivity between neurons, which achieves a significant breakthrough in the architecture formation of neural networks. In this way, the RCLSTM model exhibits a certain level of sparsity, which leads to an appealing decrease in the computational complexity and makes the RCLSTM model become more applicable in latency-stringent application scenarios. In the field of telecommunication networks, the prediction of traffic series and mobility traces could directly benefit from this improvement as we further demonstrate that the prediction accuracy of RCLSTM is comparable to that of the conventional LSTM no matter how we change the number of training samples or the length of input sequences.
\end{abstract}


\IEEEpeerreviewmaketitle

\section{Introduction}
The analysis and prediction of time series has always been the key technique in an array of practical problems, including weather forecasting, transportation planning, traffic management, and so on. 
In the domain of telecommunications, intelligent mechanisms have already been designed to track and analyze a large number of time-dependent events, such as data traffic, user location, channel load, and service requests, to mention a few \citep{7886994}. 
On the other hand, with the explosive proliferation of mobile terminals as well as the expansion of mobile Internet, the Internet of Things (IoT) and cloud computing, the mobile communication network has become an indispensable social infrastructure that is bound up with people's lives and various areas of society \citep{7886994}. However, it remains a challenging issue that how to guarantee the quality of service (QoS) and the quality of experience regardless of the dynamics of network traffic and user movements. One promising solution is to predict the varying pattern of data traffic and the location at which a mobile user will likely demand network service \citep{7886994}. Accordingly, network operators can reserve some network resources, and react effectively to network changes in near real time \citep{wang2017spatiotemporal}. However, since misplaced reservation of network resources and outdated predicted information will not only fail to support the desired QoS but also likely degrade the performance of the overall network, the prediction accuracy and complexity is of vital importance \citep{7780479}. 

Basically, the prediction goal of a time series $\left\{y_1, y_2, ...\right\}$ is to estimate the value at time $i$ based on its previous data $y_{i-1}, y_{i-2}, ...$. If we denote $\textbf{x}=\left\{y_{i-k}, y_{i-k+1}, ..., y_{i-1}\right\}, i=\left\{k, ..., n\right\}$, the goal aims at finding a function $f(\textbf{x})$ so that $\hat{y_i}=f(\textbf{x})$ is as close to the ground truth $y_i$ as possible. 

Time series analysis and prediction have been intensively studied for 40 years \citep{4840324}. In statistical signal processing, the Autoregressive Integrated Moving Average (ARIMA) model has been used to study time-varying processes. However, one limitation of ARIMA is its natural tendency to concentrate on the mean values of the past series data. Therefore, it remains challenging to capture a rapidly changing process \citep{hong2012application}. Support Vector Regression (SVR) has been successfully applied for time series prediction, but it also has disadvantages like the lack of structured means to determine some key parameters of the model \citep{hong2012application}.
In recent years, owing to the flexible structure, deep learning models are increasingly used in time series prediction \citep{witten2016data}. Specifically, Recurrent Neural Networks (RNNs), one of deep learning models, establish the reputation to cope with time series by recurrent neural connections. However, for any standard RNN architecture, the influence of a given input on the hidden layers and eventually on the neural network output would either decay or blow up exponentially when cycling around recurrent connections. To tackle this problem, Long Short-Term Memory (LSTM) has been revolutionarily designed by changing the structure of the hidden neurons in traditional RNN \citep{hochreiter1997long}. Today, research and applications of LSTM for time series prediction are proliferating. For example, Wang $\emph{et al.}$ \citep{wang2017spatiotemporal} used LSTM-based model to predict the next-moment traffic load in a specific geometric area and Alahi $\emph{et al.}$ \citep{7780479} predicted the motion dynamics in crowded scenes based on LSTM.

Generally, without customized hardware and software acceleration, the LSTM's computing time is proportional to the number of parameters. Given this disappointing characteristic, in this article, we present an approach to decrease the number of involved parameters, and thus put forward a new model that reduces the computational cost.
Inspired by the interesting finding that Feed Forward Neural Networks (FFNNs) with sparse neural connections have a similar or even superior performance in many experiments compared to the conventional FFNNs \citep{shafiee2016stochasticnet}, we introduce random connectivity to the conventional LSTM model, thus forming a new architecture with sparse neural connections, called the Random Connectivity Long Short-Term Memory (RCLSTM) model.

Our simulation model is a three-layer stack RCLSTM neural network with a memory cell size of 300 per layer. The simulation data comprise traffic data from G\'EANT networks---a pan-European research network \citep{Uhlig:2006:PPI:1111322.1111341}, and realistic user-trajectory data \citep{tkavcik2016neural}. The reasons to use the two different datasets are twofold. 
First, ANN-based models like LSTM are sensitive to the dataset and may yield significant performance changes on different subsets taken from the same database.
Therefore, it would be hard to draw conclusions from one single dataset for comparing different algorithms. Second, the prediction of both network traffic and user mobility is of significant importance in the design and optimization of telecommunication networks. For these reasons, we take advantage of the two different datasets. Our simulation results show that when compared to LSTM, the RCLSTM model is highly capable of traffic prediction and user-location forecasting with less than half the neural connections. Particularly, in the traffic prediction task, RCLSTM with even 1\% neural connections performs better than ARIMA, SVR, and FFNN, while reducing the computing time by around 30\% compared with the conventional LSTM. More interestingly, when the length of input traffic sequences increases from 50 to 500, the prediction error of RCLSTM remains basically unchanged, whereas that of LSTM increases by about 10\%.

\begin{figure*}[!t]
	\centering
	\includegraphics[width=.975\textwidth]{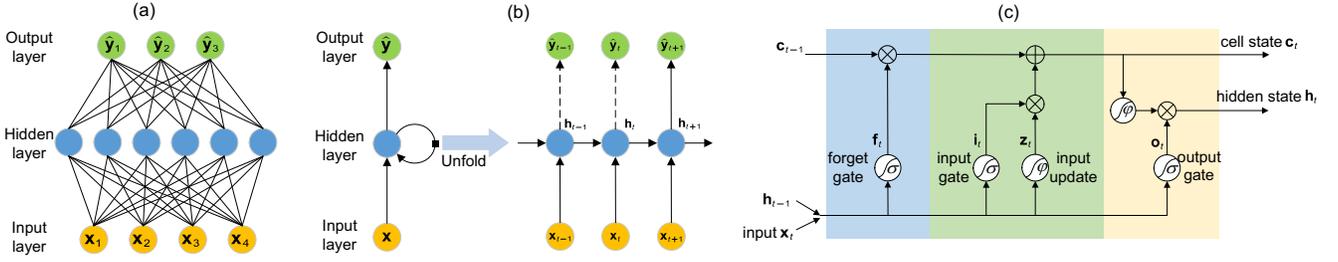}
	\captionsetup{font={scriptsize}}
	\caption{An illustration of FFNN, RNN and LSTM memory block: (a) FFNN; (b) RNN; (c) LSTM memory block.}
	\label{fig:ANNs} 
\end{figure*}

\section{An Overview of Artificial Neural Networks and LSTM}
\label{sec:OVERVIEW of ANNs}
Artificial Neural Networks (ANNs) are constructed as a class of machine learning models that can eliminate the drawbacks of the traditional learning algorithms with rule-based programming \citep{lecun2015deep}. ANNs can be classified into two main categories---FFNNs and RNNs. FFNNs usually consist of an input layer, an output layer and hidden layers (if necessary). Each layer is composed of a number of neurons and an activation function. A simple diagram of FFNNs is illustrated in Fig. \ref{fig:ANNs}(a). In FFNNs, there is no connection between the neurons within the same layer, and all neurons cannot be connected across layers,
which means the information flows in one direction, from the input layer, through the hidden layers (if any), to the output layer. FFNNs are widely used in various fields like data classification, object recognition, and image processing. However, constrained by their internal structure, FFNNs are unsuitable for handling historical dependencies.

RNNs, as another type of ANNs, are similar to FFNNs in the structure of neural layers, but allow the connections between the neurons within the same hidden layer. An illustration of RNNs can be observed in the left-hand side of Fig. \ref{fig:ANNs}(b). In addition, the right-hand side of Fig. \ref{fig:ANNs}(b) is the expanded form of the RNN model, indicating that RNNs calculate the output of the current moment from the input of the current moment $\textbf{x}_t$ and the hidden state of the previous moment $\textbf{h}_{t-1}$. Therefore, RNNs allow historical input information to be stored in the network's internal state, and are thereby capable of mapping all of the historical input data to the final output. Theoretically, RNNs are competent to handle such long-range dependencies. However, in practice, RNNs seem unable to accomplish the task. This phenomenon has been explored in depth by Hochreiter and Schmidhuber \citep{hochreiter1997long}, who explained some pretty fundamental reasons why such learning might be difficult.

\begin{figure*}[!t]
	\centering
	\includegraphics[width=.975\textwidth]{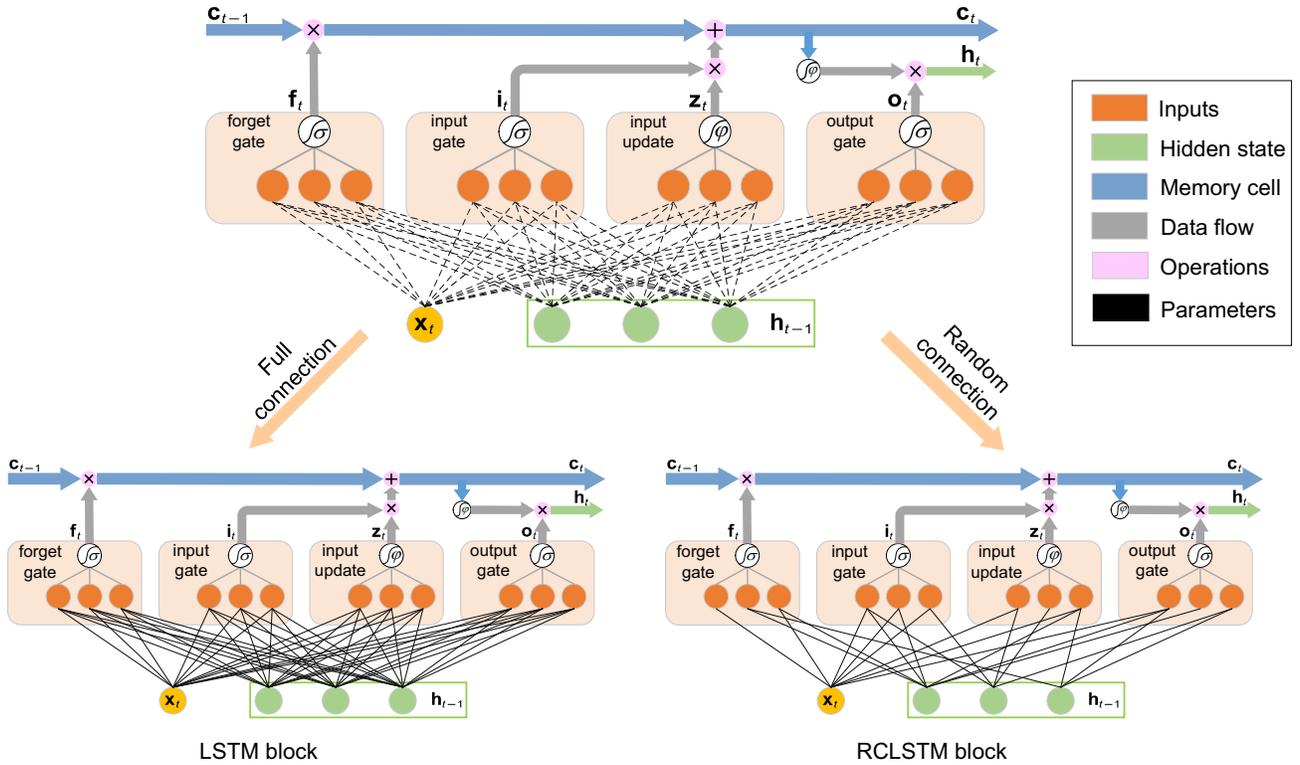}
	\captionsetup{font={scriptsize}}
	\caption{The comparison between LSTM and RCLSTM in the view of generating process of neural connections.}
	\label{fig:RCLSTM}
\end{figure*}

Long Short-Term Memory networks, usually just called ``LSTMs", are a special RNNs that are suitable for learning long-term dependencies \citep{hochreiter1997long}. The key part that enhances LSTMs' capability to model long-term dependencies is a component called memory block \citep{hochreiter1997long}. As illustrated in Fig. \ref{fig:ANNs}(c), the memory block is a recurrently connected subnet that contains functional modules called the memory cell and gates. The memory cell is in charge of remembering the temporal state of the neural network and the gates formed by multiplicative units are responsible for controlling the pattern of information flow. According to the corresponding practical functionalities, these gates are classified as input gates, output gates and forget gates. Input gates control how much new information flows into the memory cell, while forget gates control how much information of the memory cell still remains in the current memory cell through recurrent connection, and output gates control how much information is used to compute the output activation of the memory block and further flows into the rest of the neural network. Before going through the details of LSTM, some simple yet useful activation functions need to be reviewed.
The sigmoid function $\sigma(\emph{x})=\frac{1}{1+\emph{e}^{-\emph{x}}}$ and the tanh function $\varphi(\emph{x})=2\sigma(2\emph{x})-1$ are commonly used as the activation function in ANNs. 
The domain of both functions is the real number field, but the return value for the sigmoid function ranges from 0 to 1, while the tanh function ranges from -1 to 1. Fig. \ref{fig:ANNs}(c) explains in detail how LSTM works. The first step is to decide what kind of information will be removed from the memory cell state, which is implemented by a sigmoid layer (i.e., the forget gate). The next step is to decide what new information will be stored in the memory cell state. This operation can be divided into two steps. First, a sigmoid layer (i.e., the input gate) determines what will be updated, and a tanh layer creates a vector of new candidate values $\textbf{z}_t$ that can be added to the memory cell state, where the subscript $t$ denotes the current moment. Next, these two parts are combined to trigger an update to the memory cell state. To update the old memory cell state $\textbf{c}_{t-1}$ into the new memory cell state $\textbf{c}_{t}$, we can first multiply the corresponding elements of $\textbf{c}_{t-1}$ and the output of forget gate layer (i.e. $\textbf{f}_t$), which is just like the oblivion mechanism in the human brain, and then add $\textbf{i}_t*\textbf{z}_t$, where $\textbf{i}_t$ denotes the output of input gate and $*$ denotes element-wise multiplication. The last step is to decide what to output, which is realized by element-wise multiplication between the value obtained from a tanh function of $\textbf{c}_t$ and the output of a sigmoid layer (i.e., the output gate), $\textbf{o}_t$. Through the cooperation between the memory cell and the gates, LSTM is endowed with a powerful ability to predict time series with long-term dependences.

Since the invention of LSTM, a number of scholars have proposed several improvements with respect to its original architecture. Greff $\emph{et al.}$ \citep{greff2017lstm} evaluated the aforementioned conventional LSTM and eight different variants thereof (e.g., Gated Recurrent Unit (GRU) \citep{DBLP:journals/corr/ChungGCB14}) on three benchmark problems---TIMIT, IAM Online and JSB Chorales. Each variant differs from the conventional LSTM by a single and simple change. They found that the conventional LSTM architecture performs well on the three datasets, and none of the eight investigated modifications significantly improve the performance.


\section{Random Connectivity for LSTM}   
\label{sec:RCLSTM}      

The conventional LSTM (including its variants) follows the classical pattern that the neurons in each memory block are fully connected and this connectivity cannot be changed manually.
However, it has been found that for certain functional connectivity in neural microcircuits, random topology formation of synapses plays a key role and can provide a sufficient foundation for specific functional connectivity
to emerge in local neural microcircuits \citep{hill2012statistical}. This discovery is different from the conventional cases where neural connectivity is considered to be more heuristic so that neurons need to be connected in a more fully organized manner. It raises a fundamental question as to whether a strategy of forming more random neural connectivity, like in the human brain, might yield potential benefits to LSTM's performance and efficiency. With this conjecture, we built up the RCLSTM model.

In RCLSTM, neurons are randomly connected rather than being fully connected as in LSTM. Actually, the trainable parameters in LSTM only exist between the input part---the combination of the input of the current moment (i.e. $\textbf{x}_t$) and the output of the previous moment (i.e. $\textbf{h}_{t-1}$), and the functional part---the combination of the gate layers and the input update layer. Therefore, the LSTM architecture can be further depicted in Fig. \ref{fig:RCLSTM}. In our approach, whether the LSTM neurons are connected or not can be determined by certain randomness. Therefore, we use dashed lines to denote that the neural connections can be added or omitted, as depicted in the upper part of Fig. \ref{fig:RCLSTM}. If the neurons are fully connected, then it becomes a standard LSTM model. On the other hand, if the neurons are randomly connected according to some rules (which are covered in detail below), then an RCLSTM model is created. The lower right part of Fig. \ref{fig:RCLSTM} shows an example RCLSTM structure in which the neural connections are randomly sparse, unlike the LSTM model. The fundamental difference between RCLSTM and LSTM is illustrated in Fig. \ref{fig:RCLSTM}, so let us move to the implementation strategy of randomly connecting neurons. 

First, we attach a probability value to each pair of neurons that are connected by a dashed line in the upper part of Fig. \ref{fig:RCLSTM}. The probability values can obey arbitrary statistical distributions, and we choose uniform distribution in our simulations given its computational efficiency. The probability value indicates the tendency that the corresponding pair of neurons will be connected. Then we assume all neurons are connected with the same probability and carefully set a threshold to determine the percentage of connected neurons. If the probability values are greater than the threshold, the corresponding pairs of neurons are connected. Otherwise, they are prohibited from being connected. This process can be visualized as turning dashed lines into solid lines, as shown in the right-hand transformation of Fig. \ref{fig:RCLSTM}. Therefore, the RCLSTM structure can create some sparsity, considerably decreasing the total number of involved parameters to be trained and reducing the computational loads of the whole RCLSTM network.

\section{Numerical Simulations for Traffic and Mobility Prediction} 
\label{sec:SIMULATION}
\begin{figure}[!t]
	\centering
	\includegraphics[width=0.45\textwidth]{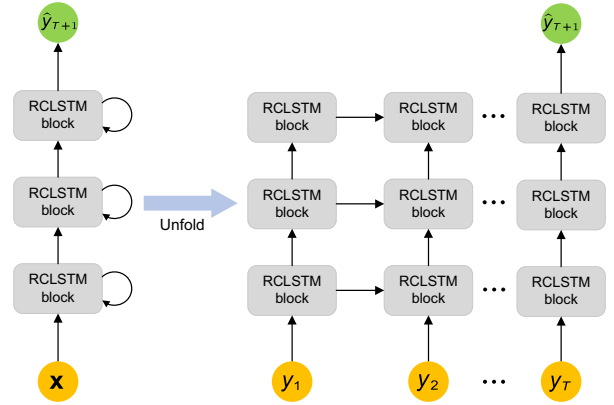}
	\captionsetup{font={scriptsize}}
	\caption{The designed RCLSTM network for simulations.}
	\label{fig：model}
\end{figure}

In this section, we focus on verifying the performance of the proposed RCLSTM model on traffic prediction and user-location forecasting. In particular, we construct a three-layer RNN whose recurrent neurons are all the newly designed RCLSTM memory blocks (for the sake of simplicity, this model is called the RCLSTM network in the following statement). The corresponding expanded form is described in Fig. \ref{fig：model}. Because the purpose of our simulations is to predict the value at the next moment given some historical data, the input data are from $y_1$ to $y_T$ where $T$ denotes the length of the input sequences, and the output of the RCLSTM network is a prediction of the actual value at the next moment denoted as $\hat{y}_{T+1}$. First of all, we take advantage of the RCLSTM network to predict traffic and user mobility, particularly compare the prediction accuracy of the RCLSTM network with that of other algorithms or models. 
Then, we adjust the number of training data samples and the length of input sequences to investigate the influence of these factors on the prediction accuracy of the RCLSTM network.

\subsection{Data Description and Processing}
\label{sec:Data}
We evaluate the model's performance on traffic prediction depending upon real traffic data from a link in the G\'EANT backbone networks \citep{Uhlig:2006:PPI:1111322.1111341}. G\'EANT is a pan-European data source for the research and education communities. These traffic data are sampled every 15 minute during a 4-month period and the unit for data points is Kbps. In this study, we selected 10772 traffic data points from 2005-01-01 00:00AM to 2005-04-30 00:00AM. The raw data are so uneven in numerical size that there are about three orders of magnitude difference between the maximum and minimum. To reduce the numerical unevenness, we first take the logarithm to base 10 of the raw data, then carry out a normalization process according to $\dfrac{\emph{\textbf{x}}-min(\emph{\textbf{x}})}{max(\emph{\textbf{x}})-min(\emph{\textbf{x}})}$, where $\emph{\textbf{x}}$ is the vector after taking the logarithm of the raw data, $min(\emph{\textbf{x}})$  and $max(\emph{\textbf{x}})$ denote the minimum and maximum value of $\emph{\textbf{x}}$, respectively. Through this process, the raw data are limited to a range between 0 and 1, which makes the training phase of ANN-based models converge faster and effectively avoid the bad local optimal solution \citep{lecun2015deep}. Real-time prediction of data traffic requires continuous data input and learning. Hence, we introduce a notion of sliding window, which indicates a fixed number of previous timeslots to learn and then predict the current data traffic. Finally, we split the processed data into two sets (i.e., a training set and a test set). The training set is used to train the RCLSTM network, and the test set is used to evaluate its prediction accuracy.

The other dataset to evaluate the capability of the RCLSTM model comes from reference \citep{tkavcik2016neural}, which contains the location history of several mobile users together with manually defined important places for every person, and is thus used for user-mobility prediction. The data for every person are taken from the Android location history service, and mobile users are asked to mark the important places that are further constructed as polygons on the map. The dataset contains four attributes, i.e., date-time, latitude, longitude, and assigned location ID, within which the last item is used to map the user's location to one of the geographical polygons. These data are collected in 1-hour intervals from 2015-08-06 to 2015-11-04. In this article, we use location data of five users. First, we encode the location ID to one--hot code. For example, suppose that in User A's trajectory, the number of different location IDs marked by User A is $m$. Then we map these different location IDs to the Numbers 1 to $m$, and use them as new location IDs. If the newly current location ID is $i$, we construct an \emph{m}-dimensional vector where only the element with index $i$ is one and the others are all zero. After transforming all the raw data to one-hot vectors, we use the sliding window to slice the processed data and finally split them into training set and test set, which follows the same procedure as mentioned earlier in the traffic data processing.

\subsection{Evaluation Metrics}
\label{sec:Evaluation}
To evaluate the performance of our proposed RCLSTM model on traffic prediction, Root Mean Square Error (RMSE) is applied to estimate the prediction accuracy. RMSE measures the squared root of the mean of the deviation squares, which quantifies the difference between the predicted values and the actual ones. The RMSE can be expressed as RMSE=$\sqrt{\frac{1}{N}\sum_{i=1}^{N}(y_i-\hat{y_i})^2}$, where $y_i$ is the actual value, $\hat{y_i}$ is the predicted value and $N$ represents the total number of predicted traffic data.

On the other hand, the evaluation metric for human mobility prediction is the accuracy level, which indicates the percentage of the correct location predictions. In our study, the accuracy level is defined as Acc=$\frac{\sum_{i=1}^{N}\mathbf{1}(y_i=\hat{y_i})}{N}$, where $y_i$ is the actual value, $\hat{y_i}$ is the predicted value, $N$ represents the total number of predicted locations and $\mathbf{1}(y_i=\hat{y_i})$ is the indicator function that equals to 1 if $y_i=\hat{y_i}$ or 0 if $y_i\ne\hat{y_i}$.

\subsection{Testing Results and Analyses}
\label{sec:results}

\subsubsection{Traffic Prediction}
\label{sec:traffic}
\begin{figure*}[!t]
	\centering
	\includegraphics[width=.975\textwidth]{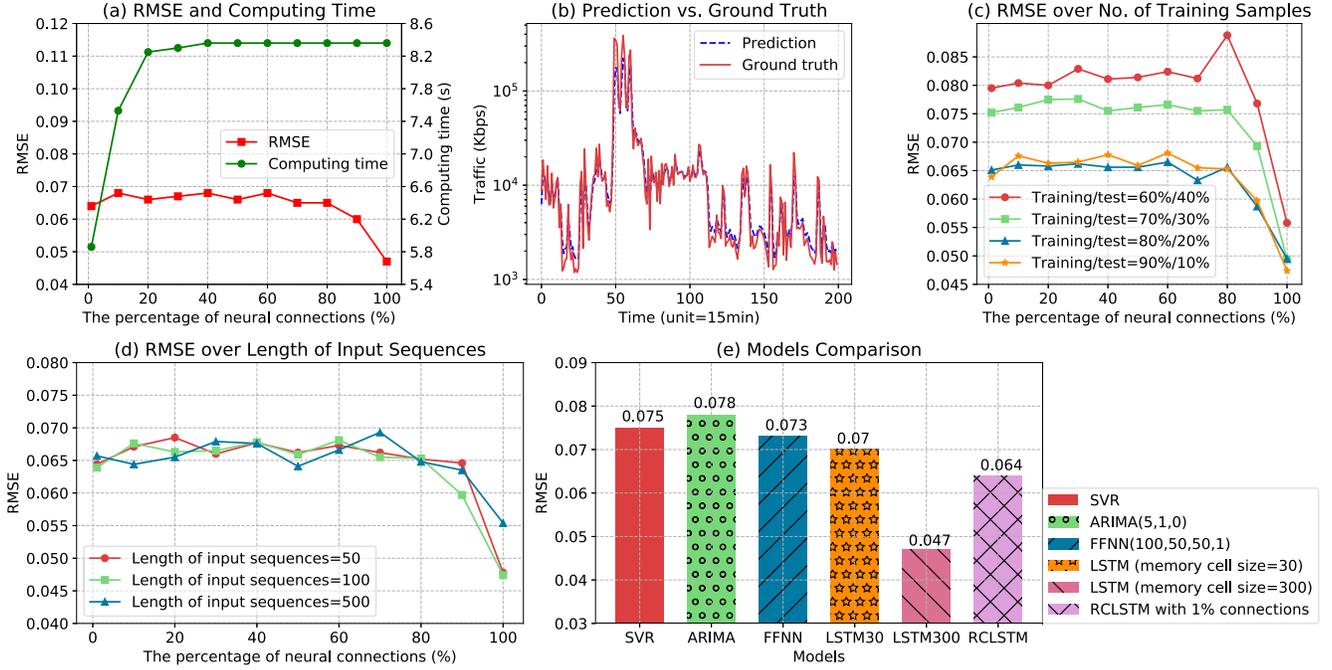}
	\captionsetup{font={scriptsize}}
	\caption{Results of traffic prediction using RCLSTM network: (a) RMSE and  computing time of RCLSTMs ; (b) predicted traffic data and actual traffic data; (c) RMSE of RCLSTMs over different number of training samples; (d) RMSE of RCLSTMs over different length of input sequences; (e) comparing  RCLSTM with SVR, ARIMA, FFNN and LSTM.}
	\label{fig:traffic} 
\end{figure*}
Fig. \ref{fig:traffic}(a) reveals the RMSE and the computing time under different percentages of neural connectivity in the RCLSTM model (note that 100\% connectivity means the fully-connected LSTM model). Notably, the probability of neural connections obeys a uniform distribution between 0 and 1. In addition, the size of the RCLSTM's memory cell is set at 300, the ratio between the number of training samples and the number of test samples is set at 9:1, and the length of input traffic sequences is 100. Fig. \ref{fig:traffic}(a) shows that the RMSE of the RCLSTM model is slightly larger than that of the LSTM model, but the RCLSTM with very sparse neural connections (i.e. 1\%) reduces the computing time by around 30\% compared with the baseline LSTM. In addition, the computing time almost stops increasing when the percentage of neural connections is larger than 20\%, which is because the method we use for accelerating calculation only works efficiently on extremely sparse matrices. Fig. \ref{fig:traffic}(b) intuitively illustrates the actual and predicted traffic values. It can be observed from the subfigure that the predicted values can match the variation trend and features of the actual values very well. Therefore, the simulation results indicate that the RCLSTM model can yield acceptable prediction capability, and effectively decrease the computational loads and complexity.

Then we investigate how the predictive capability of the RCLSTM model is influenced by the number of training samples and the length of input traffic sequences. 
First we train the models with 90\%, 80\%, 70\% and 60\% of the processed data, respectively, while fixing both the size of the memory cell (set at 300) and the length of the input sequences (set at 100), and test the trained models with the remaining data. Then we vary the length of the input sequences from 50 to 500 while keeping the values of the other hyperparameters fixed. The simulation results are shown in Fig. \ref{fig:traffic}(c) and (d). It can be observed from Fig. \ref{fig:traffic}(c) that RCLSTM models are more sensitive to the number of training samples than LSTM, because when the number of training samples increases, the RMSEs of the RCLSTM models vary more significantly than that of the LSTM model. Fig. \ref{fig:traffic}(d) gives the related results and shows that RCLSTM models are less susceptive to the length of the input sequences than LSTM model.

\begin{figure*}[!t]
	\centering
	\includegraphics[width=.975\textwidth]{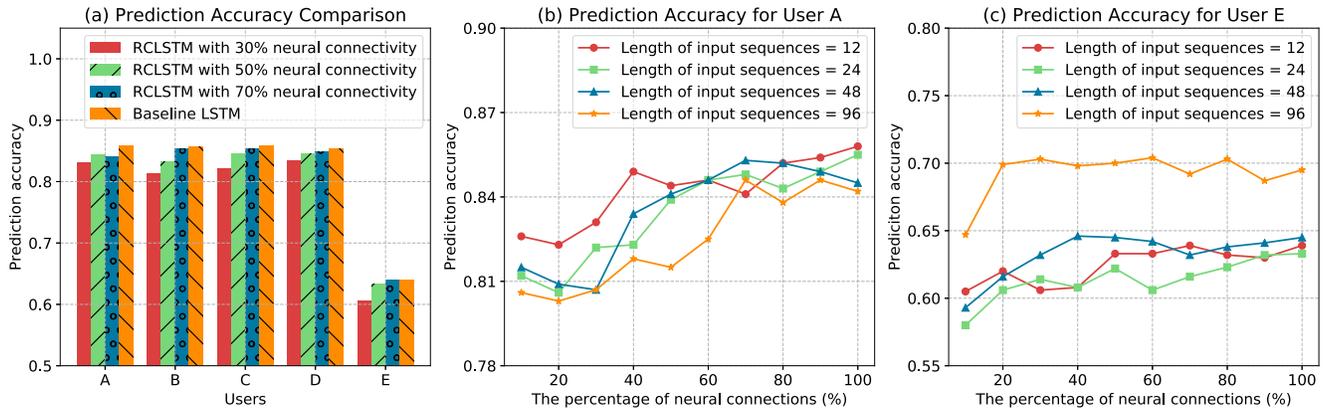}
	\captionsetup{font={scriptsize}}
	\caption{Results of human mobility prediction using RCLSTM: a) prediction accuracy of RCLSTMs for different users; b) prediction accuracy of RCLSTMs over different length of input sequences for User A; c) performance of RCLSTMs over different length of input sequences for User E.}
	\label{fig:mobility} 
\end{figure*}

Finally, we compare the RCLSTM with three well-known prediction techniques---SVR, ARIMA, and FFNN. The hyper-parameters of these algorithms are as follows:
\begin{itemize}
	\item SVR: The number of input features is 100, the kernel is a Radial Basis Function (RBF) and the tolerance for the stopping criterion is 0.001.
 	\item ARIMA($p$, $d$, $q$): The number of autoregressive terms (i.e. $p$) is 5, the number of nonseasonal differences needed for stationarity (i.e. $d$) is 1, and the number of lagged forecast errors in the prediction equation (i.e. $q$) is 0.
	\item FFNN: The number of input features is 100 and the numbers of neurons in both the first hidden layer and the second hidden layer are 50.
\end{itemize}

In addition, since the LSTM with a memory cell size of 30 has almost as many trainable parameters as the RCLSTM with a memory cell size of 300 and 1\% neural connections, we put it into the comparison list as well. The simulation results are shown in Fig. \ref{fig:traffic}(e), which reveals that LSTM with a memory cell size of 300 performs much better than the others, followed by the RCLSTM with the memory cell size of 300 and 1\% neural connections. Interestingly, the RCLSTM model performs better than the LSTM with the memory cell size of 30, which is probably due to a degree of overfitting that exists in the latter \citep{lecun2015deep}.

\subsubsection{Human Mobility Prediction}
\label{sec:mobility}
The results of user-mobility prediction with the RCLSTM model are shown in Fig. \ref{fig:mobility}. Fig. \ref{fig:mobility}(a) shows the respective prediction accuracy for the five users with the RCLSTM model, where the probability of neural connections obeys a uniform distribution between 0 and 1. In addition, the size of the memory cell is 150, the length of input sequences is 12, and the training samples account for 90\% of the processed data. There is a slight difference in the prediction results for different users. For example, Users A and D are both university students with a part-time job, and thus they almost follow the same behavioral pattern in school or work, which results in highly expected predictability. On the other hand, User E is running his own business and is more likely to travel a lot with unfixed schedule, consequently having low expected predictability \citep{tkavcik2016neural}. Although the prediction accuracy of the RCLSTM model is not as good as that of the LSTM model, the RCLSTM model with high sparsity of neural connections can compute faster than the LSTM, similar to the traffic prediction scenario.

In order to further investigate the capability of the RCLSTM model to characterize long-term dependencies of user mobility, we carry out simulations with the different length of input sequence data for Users A and E, consistent with the scenario of traffic prediction. The results are shown in Fig. \ref{fig:mobility}(b) and (c), respectively. It can be observed from the subfigures that the RCLSTM models can catch up with LSTM in terms of prediction accuracy, regardless of the length of the input sequences. Remarkably, for User A, increasing the length of input sequences scarcely affects the prediction accuracy, whereas for User E, the prediction accuracy improves when the length of the input sequences increases. From the perspective of data analysis, the mobility data of User E is more irregular than that of User A, so both the RCLSTM model and the baseline LSTM model need more input data to find the regular movement patterns of User E.

\begin{remark}
	The RCLSTM model shows promise for manifesting strong traffic and user-mobility prediction capabilities while reduces the number of parameters to be trained, which in effect decreases the computational load and complexity.
\end{remark}

\section{Conclusion}
\label{sec:conclusion}
In this article, we have addressed the importance of leveraging deep learning for time series prediction. In particular, we have reinvestigated the issues of traffic prediction and user-mobility forecasting with deep learning and proposed a new model named RCLSTM by revolutionarily redesigning the conventional LSTM model. The basic idea behind the RCLSTM model is to construct neural networks by forming and realizing a random sparse graph. 
We have checked the effectiveness of the RCLSTM model by predicting the dynamics of traffic and user locations through various temporal scales. In traffic prediction, we have demonstrated that the RCLSTM model with 1\% neural connections reduces the computing complexity by 30\% compared with the standard LSTM model. Although the characteristic of sparse neural connections may cause a degradation of approximately 25\% in prediction accuracy, the RCLSTM model still outperforms SVR, ARIMA, FFNN, even the LSTM model with the same number of parameters. In addition, along with the adjustment of the length of input sequences and the number of training samples, the RCLSTM models fluctuate in line with the LSTM model in terms of prediction accuracy. In summary, it can be expected the RCLSTM model with lower computing costs and satisfactory performance will play an essential role in time series prediction in the future intelligent telecommunication networks.

\renewcommand\refname{\textsc{Reference}}
\small

\bibliographystyle{IEEEtran}
\bibliography{rclstm.bib}

\end{document}